\documentclass{article}
\usepackage{spconf,amsmath,graphicx}
\usepackage{booktabs}
\usepackage{epstopdf}

\title{Transfer Learning for Input Estimation of Vehicle Systems}

\name{Liam~M.~Cronin$^{\star}$ \qquad Soheil~Sadeghi~Eshkevari$^{\star \dagger}$ \qquad Debarshi~Sen$^{\star}$ \qquad Shamim~N.~Pakzad$^{\star}$ \thanks{© 2020 IEEE. Personal use of this material is permitted. Permission from IEEE must be obtained for all other uses, in any current or future media, including reprinting/republishing this material for advertising or promotional purposes, creating new collective works, for resale or redistribution to servers or lists, or reuse of any copyrighted component of this work in other works.}}

\address{$^{\star}$ Department of Civil and Environmental Engineering, Lehigh University, Bethlehem, PA 18015 USA \\ $^{\dagger}$ Senseable City Lab, Massachusetts Institute of Technology, Cambridge, MA 02143 USA}

\begin{document}
%
\maketitle
\begin{abstract}
This study proposes a learning-based method with domain adaptability for input estimation of vehicle suspension systems. In a crowdsensing setting for bridge health monitoring, vehicles carry sensors to collect samples of the bridge's dynamic response. The primary challenge is in preprocessing; signals are highly contaminated from road profile roughness and vehicle suspension dynamics. Additionally, signals are collected from a diverse set of vehicles vitiating model-based approaches. In our data-driven approach, two autoencoders for the cabin signal and the tire-level signal are constrained to force the separation of the tire-level input from the suspension system in the latent state representation. From the extracted features, we estimate the tire-level signal and determine the vehicle class with high accuracy ($98\%$ classification accuracy). Compared to existing solutions for the vehicle suspension deconvolution problem, we show that the proposed methodology is robust to vehicle dynamic variations and suspension system nonlinearity.
\end{abstract}
\begin{keywords}
transfer learning, deep learning, inverse problem, structural health monitoring, mobile sensing.
\end{keywords}
\section{Introduction}
\label{sec:intro}
Deteriorating transportation infrastructure is a major public safety concern, and an efficient, real-time, and widespread network to guide decision makers for resource allocation is a necessity \cite{american20172017, willsher2018bridges}. Vibration-based monitoring with output-only system identification and damage detection algorithms is a popular approach for bridge safety management due to its low-cost and simplicity \cite{kim2007health}. However, these methods are restrictive since they require manual placement of a limited number of sensors that only capture a snapshot of the bridge health at the time of data collection. A promising candidate to remedy these issues is a mobile sensing framework employing vehicles with on-board sensors, such as smartphones, as a network, which continually supply a large quantity of inexpensive but highly contaminated data of the bridges' dynamic response. \par

Smartphones use a variety of sensors to enhance user experience. Although designed primarily for customer use, these sensors can detect secondary information about their surroundings with many potential applications for crowdsensing. Applications in civil engineering include transportation planning \cite{calabrese2010real, wang2012understanding}, road condition assessment \cite{eriksson2008pothole}, and indirect bridge health monitoring. In a comprehensive experimental study, \cite{matarazzo2020crowdsourcing} successfully extracted multiple modal frequencies of the Golden Gate bridge using merely smartphones and ride-sharing data from passing vehicles. \cite{mei2019crowdsourcing} extracted damage features for bridges based on Mel-frequency spectral coefficients using smartphone data. \par

A prevailing issue in extracting high level information from mobile sensors in a realistic scenario is the noise from collecting the response inside a moving vehicle; the road conditions, vehicle suspension system, and vehicle speed play critical roles in the signal quality \cite{yang2004extracting}. Recent studies have shown that by solving an inverse problem - deconvolving the measured cabin accelerations to recover the tire-level input - the detrimental suspension effects could be mitigated \cite{eshkevari2020bridge,yang2020measuring}. To incorporate error in modeling and measurements, \cite{nayek2020extraction} proposed a Gaussian process latent force model that successfully estimated the road input for a linear, damped, two degrees of freedom (DOF) quarter car suspension model. However, all these models require complete knowledge about the system which is less practical for real-world applications. To address this, \cite{eshkevari2020bridge} also proposed a data-driven method based on blind source separation (BSS) techniques. While avoiding the modeling complications, this method has alternate limitations from the BSS assumptions pertaining to the system linearity and signal's frequency distribution. In another data-driven approach using neural networks, our research lab at Lehigh University developed a learning-based filter that performed successfully on real vehicle suspension data, however the model is limited to the single system that it was trained for. The objective of this paper is to develop a learning-based method for tire-level response estimation that can be transferred for a wide variety of vehicles with domain adaptability while extracting physically interpretable latent information. \par

\section{Literature Review}
\label{sec:lit_review}
Transfer learning (TL) is an emerging and extremely popular domain in machine learning (ML) that deals with transferring knowledge learned for a specific task to other models with similar but not exactly the same task. In particular, the underlying assumption in ML is that the data used in training conforms with the empirical distribution which is hard to achieve and can cause inaccurate inference. One of the successful applications of TL is in domain-invariant deep neural networks that have been trained for a general task (e.g., Word Embedding \cite{reimers2020making}) based on a specific dataset (e.g., English language), then applied to other datasets with inherently different distributions (e.g. French language) by learning new embedding maps. Domain adaptation (DA) as one type of TL paradigms bridges the gap between data distributions given that the task at hand is still the same.\par

Over recent years, many approaches have been taken for the DA problem. \cite{pan2008transfer} used DA for estimating locations of users on a WiFi network. \cite{donahue2014decaf} analyzed a convolutional neural network's (CNN) ability to extract features that are generalizable for alternative tasks, and showed that shallow features are more general while deep layers extract more task-specific characteristics. \cite{tzeng2014deep} showed that by binding extracted features between two similar domains during training, a domain-insensitive feature space can be constructed through which the network converts the conditional distributions of the input data to an unconditional distribution. Despite its insightful solution, this method requires to be combined with a robust and reliable feature extractor network. Autoencoders (AEs) are good candidates for this goal.\par

\begin{figure}[htb]
    \centering
    \includegraphics[width=60mm]{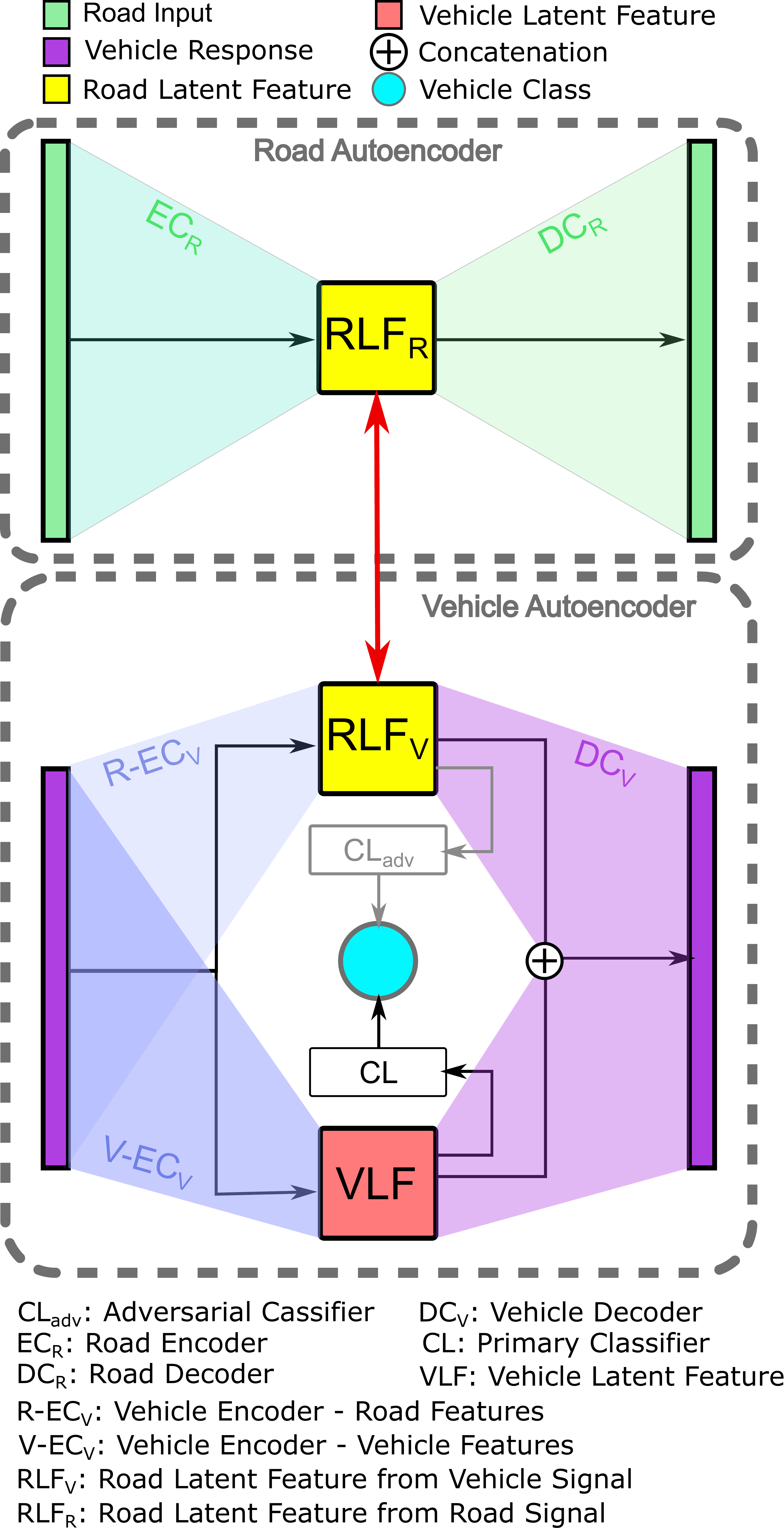}
    \caption{Schematic architecture of the proposed approach}
    \label{fig:schematic}
\end{figure}

Autoencoders are unsupervised neural networks for nonlinear dimensionality reduction and feature extraction which inherently learn the underlying low-dimensional latent information and have been successful for noise canceling and generative applications \cite{bengio2013representation}. AEs consist of two primary parts: (1) an encoder that transforms the input to an arbitrary latent space that is most representative of the underlying distribution and (2) a decoder that attempts to reconstruct the original input from the latent feature. However, basic AEs do not always create robust features. To aid this shortcoming, noise is added to the input or the latent features or alternatively, probabilistic encoders are introduced to construct distributions for the latent features (i.e., denoising AEs and variational AEs) \cite{doersch2016tutorial, vincent2008extracting}. These training approaches create a smooth latent state representation meaning similar inputs have similar latent representations, allowing for generative applications through latent state interpolations. \par




In this research, we aim to design an autoencoder that maps a signal collected in the cabin of a vehicle into two latent features (vehicle-specific and road-specific) and then tie the road-specific feature space to the one extracted from the road autoencoder. By doing so, one can estimate the vehicle input signal by cross-inferring the two autoencoders (using the latent feature from one autoencoder as an input to the other's decoder).\par


\section{Methodology}
\label{sec:method}
\begin{figure}[htb]
    \centering
    \includegraphics[width=80mm]{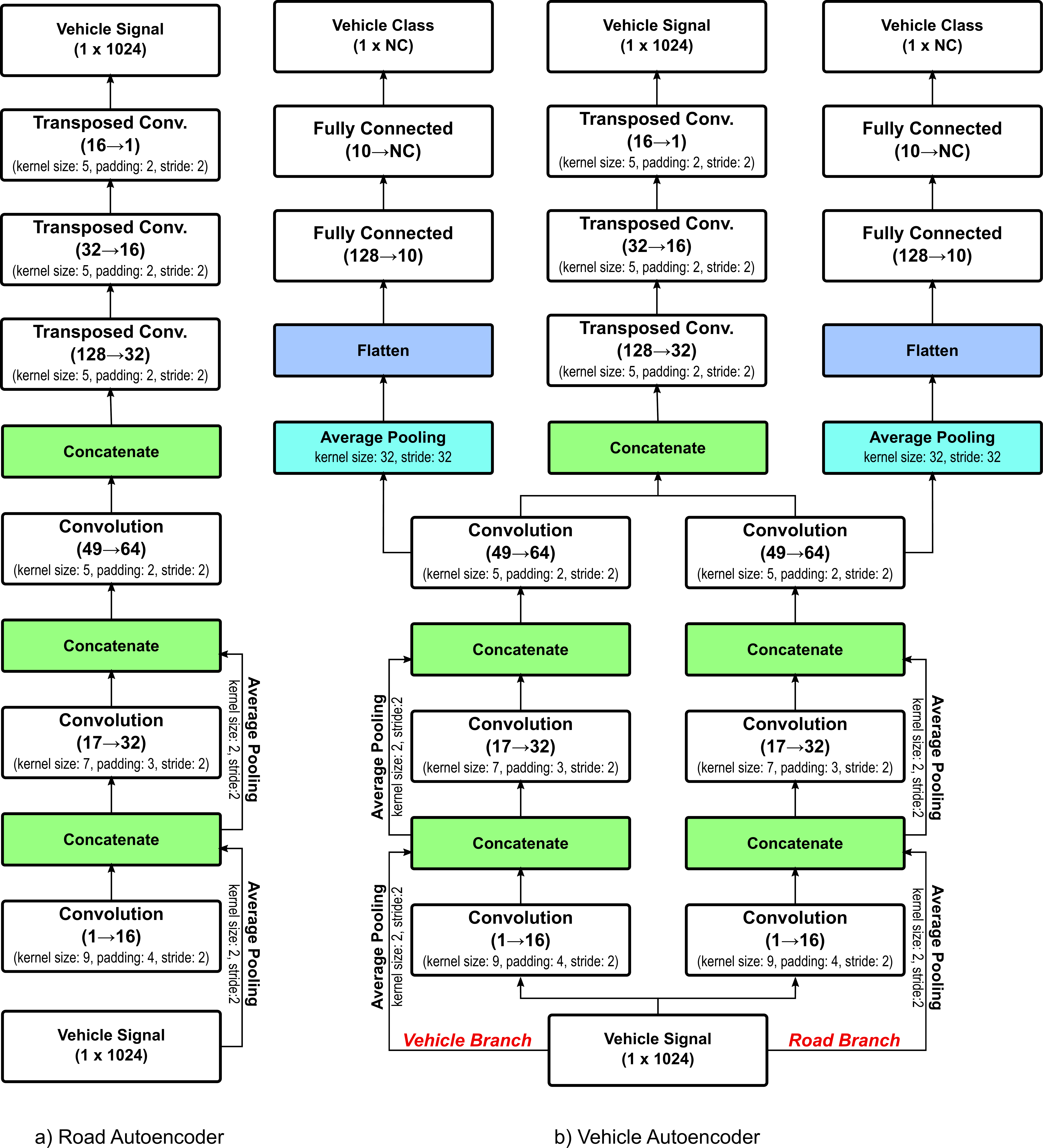}
    \caption{Network architectures for the autoencoders}
    \label{fig:net}
\end{figure}

The goal of this network is to deconvolve the signal collected from the cabin of the vehicle with respect to the suspension system to recover the system's input. Autoencoders are used to extract features such that the road (system input) latent features are void of information about the vehicle type (system). To extract these features, the network has two autoencoders for the road signal and cabin signal with the road latent feature being shared between the two networks. An adversarial training setup forces the road specific latent feature to be invariant to the vehicle type, meaning the only information left is that of the road profile. Figure \ref{fig:schematic} depicts the overall structure of the network while Figure \ref{fig:net} displays more detailed information on layer parameters. \par

The network consists of two main sections: road autoencoder and vehicle autoencoder. The road encoder ($EC_R$) transforms the road signal to an arbitrary latent space creating the road latent feature ($RLF$). Using $RLF$, the road decoder ($DC_R$) attempts to reconstruct the original road signal. The reconstruction loss for the road is the mean squared error (MSE) between the input and reconstruction ($\mathcal{L}_1$). The vehicle autoencoder also contributes a corresponding reconstruction loss, $\mathcal{L}_2$.  $RLF$ has an additional constraint from vehicle to the road encoder ($R\mbox{-}EC_V$), which tries to map the cabin signal to the same latent space. This is achieved by minimizing the difference between $RLF$ created by $EC_R$ and $R\mbox{-}EC_V$ ($\mathcal{L}_3$). Subsequently, we use $R\mbox{-}EC_V$ and $DC_R$ for estimating the vehicle input (i.e., network cross-inference). 

The primary purpose of this network is for the tire-level estimation, however, additional information can also be recovered about the type of vehicle and the expected accuracy of the network, which is directly related to the dynamics of the vehicle. An additional cabin signal encoder ($V\mbox{-}EC_V$) is used to extract vehicle information ($VLF$) from the cabin signal which is used as an input for the vehicle classifier ($CL$) and $DC_R$. $CL$, with classification loss $\mathcal{L}_5$, attempts to classify vehicle type. The cabin signal can be reconstructed by combining the road and vehicle information through concatenating $RLF$ and $VLF$ as an input to the vehicle decoder ($DC_v$). However, there is no guarantee without additional constraints that information from the car is not embedded in $RLF$ as well from this optimization process. An adversarial training process is used to apply this constraint to the latent feature. \par

The networks in Figure \ref{fig:schematic} are trained in an adversarial process to make the vehicle type indiscernible to the network in $RLF$. Using an adversarial training process, all other parameters are trained with the loss shown in Equation \ref{eq:01} while the weights of $CL_{adv}$ are not trainable. As shown in this equation, the classification loss of $CL_{adv}$, $\mathcal{L}_4$ appears with a negative sign, implying the intention for maximization. Every $K$ (user defined parameter) epochs, all parameters of the networks except for $CL_{adv}$ weights are frozen and $CL_{adv}$ parameters are optimized for $\mathcal{L}_4$ minimization. By iterating this process, the network makes the latent feature invariant to the car type and the adversarial classifier is unable to predict vehicle classes with accuracy greater than $1/n$ with $n$ being the number of cars in the dataset.\par


\begin{equation}\label{eq:01}
    \mathcal{L}_{total} = \mathcal{L}_1 + \mathcal{L}_2 + \mathcal{L}_3 - \mathcal{L}_4 + \mathcal{L}_5
\end{equation}

\begin{figure}[htb]
    \centering
    \includegraphics[width=50mm]{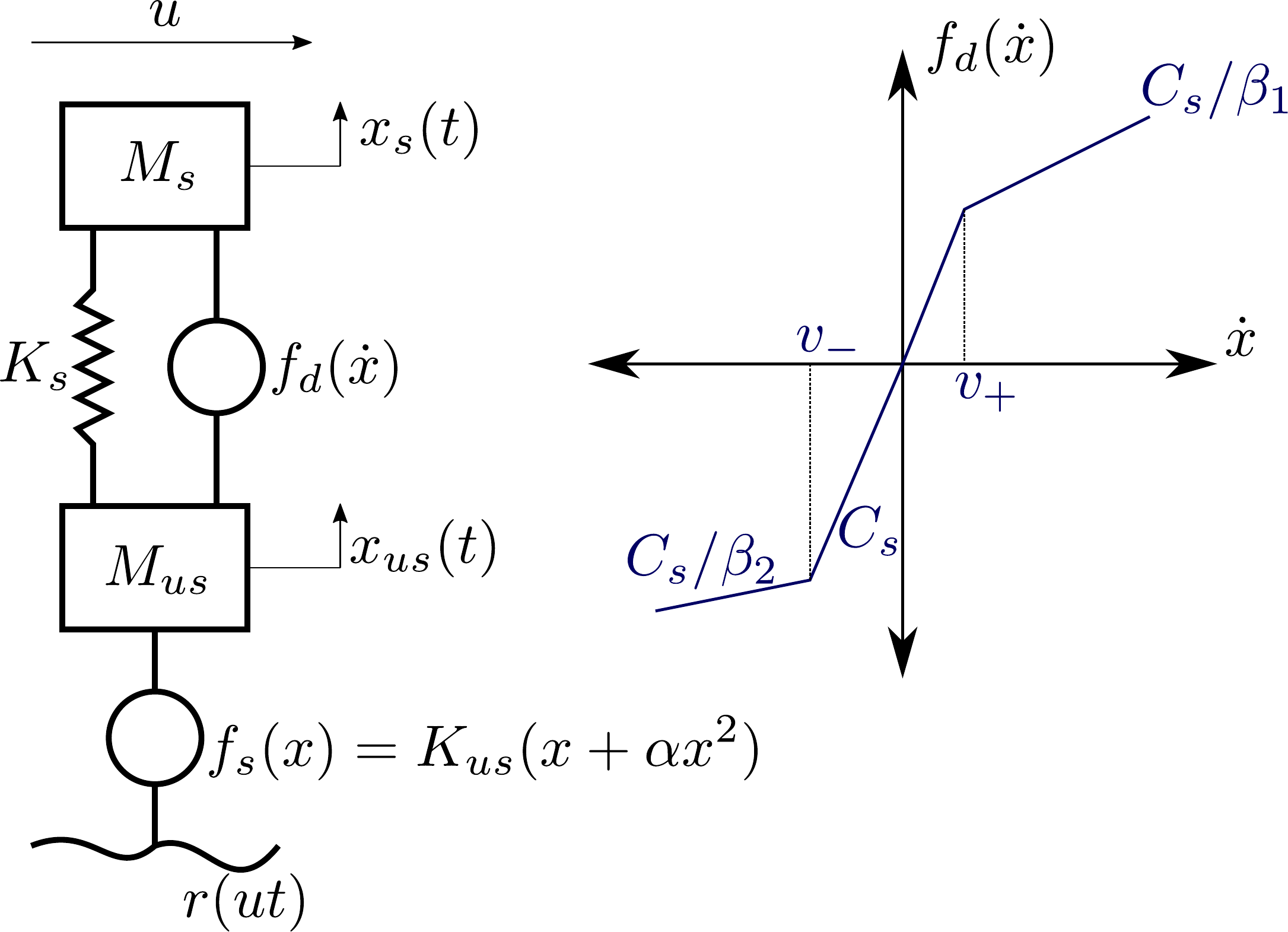}
    \caption{Nonlinear quarter car model}
    \label{fig:nonlin_vehicle}
\end{figure}

We demonstrate the efficacy of the proposed architecture through a numerical case study, wherein we train our autoencoders using response from five different vehicles modeled as nonlinear two degree of freedom (DOF) quarter cars. The nonlinearity stems from a bilinear damping of the suspension and a second order polynomial stiffness of the tire \cite{nonlin_vehicle}. Figure \ref{fig:nonlin_vehicle} shows the quarter car model along with its parameters. We generate the road profile as a stochastic field from the power spectral density $G(\lambda) = G_{\lambda_0}(\lambda/\lambda_0)^\nu$, with $\nu = -2$, $G_{\lambda_0} = \gamma*6 \times 10^{-3}$, $\gamma \sim U(0.1)$,  and $\lambda$ being the spatial frequency \cite{dodds_jsv}. Table \ref{tab:vehicle_param} lists the model parameters for each vehicle class used for training. The other parameters of the model are assumed the be constant for all the vehicles with the following values: $\beta_1 = 4.23$, $\beta_2=15.49$, $v_{+} = 0.0045$ $m/s$, $v_{-}=-0.005$ $m/s$, $\alpha=0.1$ $m^{-1}$, and $u = 5$ $m/s$.

\begin{table}[htb]
    \tiny
    \centering
    
    \begin{tabular}{c c c c c c}
    \toprule
    \textbf{Class} & $M_s$ ($kg$) & $M_{us}$ ($kg$) & $C_s$ ($N/m/s$) & $K_s$ ($N/m$) & $K_{us}$ ($N/m$) \\
    \midrule
    1 & 305.6 & 150.6 & $1.61 \times 10^3$ & $1.77 \times 10^4$ & $1.48 \times 10^5$ \\
    2 & 429.8 & 13.39 & $2.46 \times 10^3$ & $3.48 \times 10^4$ & $2.68 \times 10^5$ \\
    3 & 487.2 & 129.7 & $5.79 \times 10^3$ & $2.32 \times 10^4$ & $3.11 \times 10^5$ \\
    4 & 2141 & 74.08 & $4.34 \times 10^3$ & $1.22 \times 10^5$ & $4.93 \times 10^5$ \\
    5 & 372.9 & 206.5 & $2.76 \times 10^3$ & $1.76 \times 10^4$ & $1.69 \times 10^6$ \\
    \bottomrule
    \end{tabular}
    \caption{Vehicle model parameters for each class}
    \label{tab:vehicle_param}
\end{table}

We generate $1,000$ samples from the model including road input acceleration ($\ddot{r}(ut)$), unsprung mass ($\ddot{x}_{us}(t)$), and sprung mass ($\ddot{x}_s(t)$) acceleration signals. We split the data $7:3$ into training and evaluation datasets and subsequently train the road and vehicle autoencoders.

\section{Results}
\label{sec:resutls}
\begin{figure}[htb]
    \centering
    \includegraphics[width=65mm]{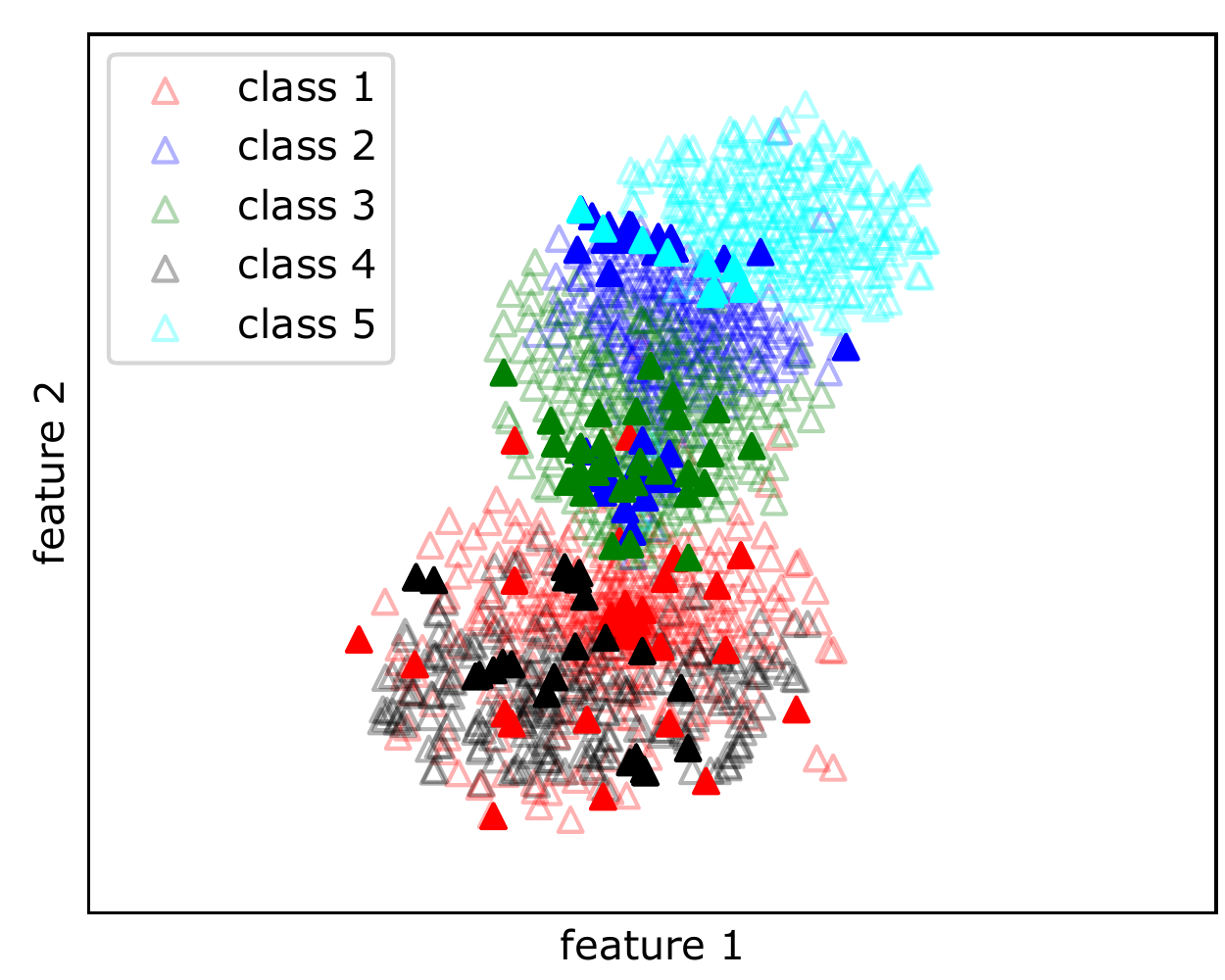}
    \caption{Two-dimensional visualization using t-SNE of $VLF$. The open and filled markers represent the five vehicles used in training and vehicles with $40\%$ perturbed mechanical properties, respectively.}
    \label{fig:tsne}
\end{figure}

\begin{figure}[htb]
    \centering
    \includegraphics[width=70mm]{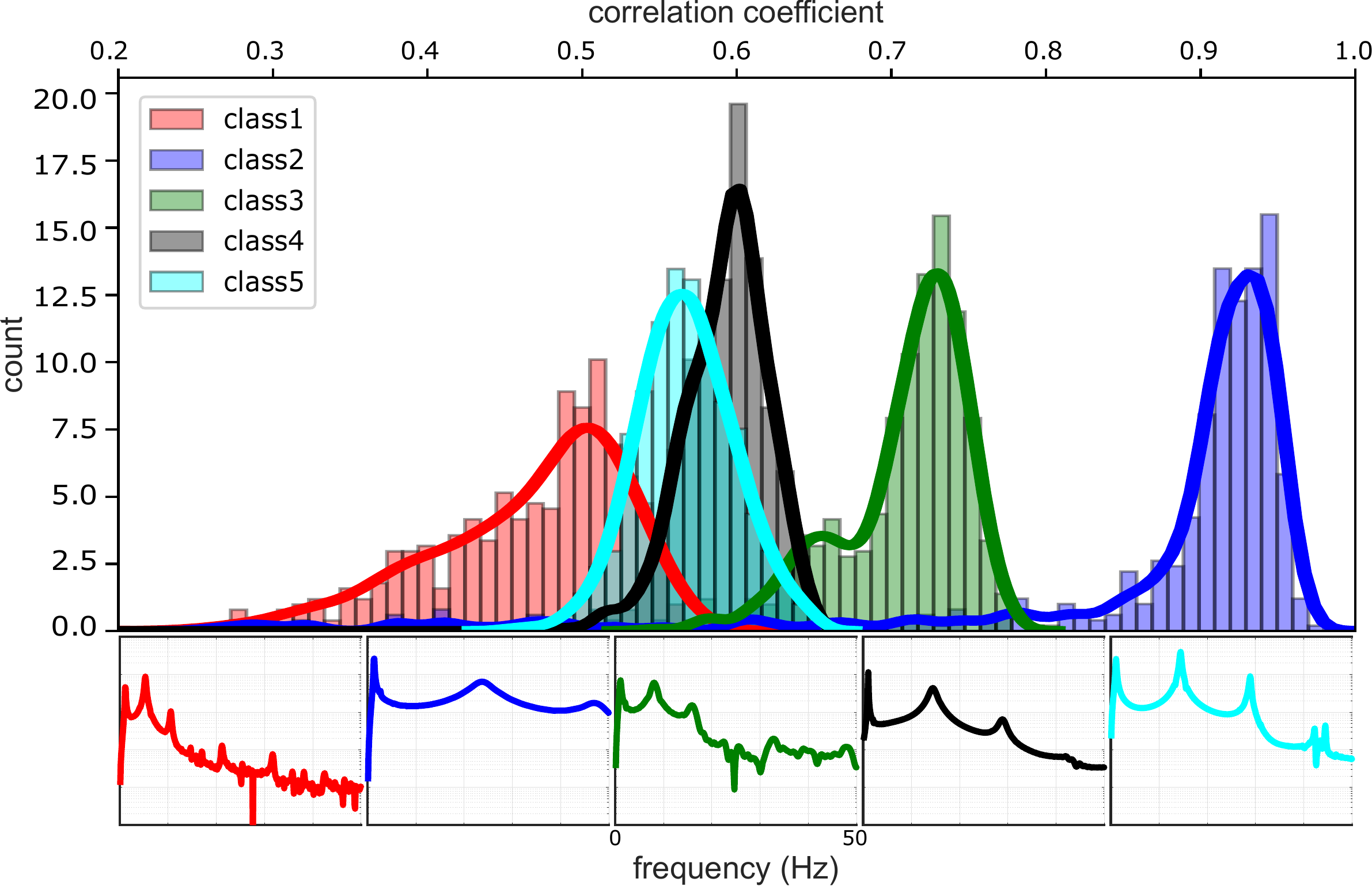}
    \caption{Top: histogram of correlation coefficients on the estimation results of $1,580$ samples from the five vehicles used in training. Bottom: nonlinear transfer functions of five vehicle classes derived by averaging the input-output relationship for a range of input impulse amplitudes.}
    \label{fig:hist}
\end{figure}

One of the goals of this network is to separate information about the road and vehicle into $RLF$ and $VLF$, respectively. Figure \ref{fig:tsne} is a two dimensional visualization of $VLF$ using t-SNE \cite{maaten2008visualizing}. Not only do the five cars used in training produce distinctive clusters, but also cars with perturbed ($1\%$, $5\%$, $10\%$, $15\%$, $20\%$, $30\%$ and $40\%$ variations) mechanical properties have similar latent representations to those most similar used in training. This confirms that the vehicle latent feature space is a continuous domain with similar cars closely spaced. In addition, having split and distinctive clusters enables $CL$ to achieve high classification accuracy ($95\%$, $93\%$, $91\%$, $86\%$, $87\%$, $76\%$, and $68\%$, respectively.). This is important since not all cars achieve the same tire-level estimation accuracy, and a robust classifier enables to cherry pick real-world samples that belong to the vehicle classes with high tire-level estimation accuracy (Figure \ref{fig:hist}).

The top portion of Figure \ref{fig:hist} depicts the tire-level estimation accuracy. The histograms were generated using the network to estimate $1,580$ samples of testing data with the correlation coefficient as a metric for accuracy. The plot shows coefficients ranging from $0.20$ up to $0.98$ for different classes. The best results are found for class 2 with median of $0.92$. Note that the tire-level estimations are made by a cross-inference of two autoencoders. In other words, the networks are not directly trained for estimating inputs from outputs and still the accuracy results are promising. Interestingly, the correlation coefficients estimates of each class of vehicles creates unimodal distributions implying that the properties of the sensing vehicle determines the accuracy of the tire-level input estimation. \par
 
The bottom portion of Figure \ref{fig:hist} shows the nonlinear transfer functions for the five vehicles used in training. Vehicles have a variety of suspension systems that affect the input signals differently. In other words, the output signal is a convolved version of the input signal with respect to these nonlinear transfer functions. Therefore, the vehicle classes that have low amplitude in a wide range of frequencies (e.g., class 1,3, and 4), and have sharp low-damping spikes in the frequency domain (e.g., class 4 and 5) filter out the significant fraction of the frequency contents in the input. As a result, the filtered signal is not informative and input retrieval task becomes nearly impossible. This can explain the lower accuracy of classes 1, 4, and 5 with respect to classes 2 and 3. We believe the reduced accuracy for input recovery in some classes is inevitable and cannot be improved due to physical limitations. However, using $CL$ (Figure \ref{fig:schematic}) we can always select samples from reliable and high accuracy classes. 

\section{Conclusion}
\label{sec:conclusion}
In this study we present a novel framework for tire-level input estimation for a variety of suspension systems through domain adaptation. We demonstrate how the network separates the road and vehicle information from the cabin signal to create two physically meaningful latent features: road ($RLF$) and vehicle ($VLF$). $RLF$ is decoded for accurate tire-level input estimation for unknown vehicle suspension systems by cross-inferring two autoencoders. It is expected that not all suspensions allow for accurate tire-level estimations since their transfer functions act as band-limited filters. In order to avoid these physically restrictive cases, $VLF$ is used to classify the type of vehicle and determine if the inference will be viable. In addition, we showed that our training process leads to distinct vehicle clusters in the vehicle latent space in an unsupervised manner. The proposed approach enables infrastructure health monitoring using crowdsourced data from traffic networks which is robust to variations in vehicle systems.

\vfill\pagebreak

\bibliographystyle{IEEEbib}
\bibliography{strings,refs}

\end{document}